\pdfoutput=1
\documentclass[11pt]{article}
\usepackage[final]{acl}
\usepackage{times}
\usepackage{latexsym}
\usepackage[T1]{fontenc}
\usepackage[utf8]{inputenc}
\usepackage{microtype}
\usepackage{inconsolata}
\usepackage{graphicx}

\usepackage{xspace,mfirstuc,tabulary}
\usepackage[utf8]{inputenc}
\usepackage[most]{tcolorbox}
\usepackage{pifont} 
\usepackage{todonotes} 
\usepackage{bbding}
\usepackage{soul} 
\usepackage{xcolor,colortbl} 
\usepackage{appendix} 
\usepackage{enumitem} 
\usepackage{booktabs} 
\usepackage{algorithm}
\usepackage{algpseudocode}
\usepackage{multirow} 
\usepackage{array} 
\usepackage{arydshln} 
\usepackage{amsmath, amssymb} \usepackage{algpseudocode} 
\usepackage{hyperref}
\usepackage[table]{xcolor}
\usepackage[normalem]{ulem}
\definecolor{ForestGreen}{rgb}{0.13, 0.55, 0.13}
\definecolor{darkred}{RGB}{139, 0, 0}
\usepackage{fontawesome5} 

\title{\textit{Measuring What Matters!!} Assessing Therapeutic Principles in Mental-Health Conversation}

\author{Abdullah Mazhar$^\dagger$, Het Riteshkumar Shah$^\dagger$, Aseem Srivastava$^\ddagger$\textsuperscript{,}\thanks{Work done while at IIIT Delhi.}, \\ \textbf{Smriti Joshi$^\sharp$}, \textbf{Md. Shad Akhtar$^\dagger$}\\
  $^\dagger$IIIT Delhi; $^\ddagger$MBZUAI; $^\sharp$Wysa\\
  \texttt{\{abdullahm, het22213, shad.akhtar\}@iiitd.ac.in,}\\ 
  \texttt{smriti@touchkin.com, aseem.srivastava@mbzuai.ac.ae} \\
}

\newcommand{\model}{\texttt{CARE}}
\newcommand{\dataset}{\texttt{FAITH-M}}
\newcommand{\mm}[1]{\textcolor{black}{#1}}

\begin{document}
\maketitle

\begin{abstract}
The increasing use of large language models in mental health applications calls for principled evaluation frameworks that assess alignment with psychotherapeutic best practices beyond surface-level fluency. While recent systems exhibit conversational competence, they lack structured mechanisms to evaluate adherence to core therapeutic principles. In this paper, we study the problem of evaluating \emph{AI-generated therapist-like responses} for clinically grounded appropriateness and effectiveness. We assess each therapists utterance along six therapeutic principles: \textit{non-judgmental acceptance}, \textit{warmth}, \textit{respect for autonomy}, \textit{active listening}, \textit{reflective understanding}, and \textit{situational appropriateness} using a fine-grained ordinal scale. We introduce \dataset, a benchmark annotated with expert-assigned ordinal ratings, and propose \model, a multi-stage evaluation framework that integrates intra-dialogue context, contrastive exemplar retrieval, and knowledge-distilled chain-of-thought reasoning. Experiments show that \model\ achieves an F-1 score of 63.34 versus the strong baseline Qwen3 F-1 score of 38.56 which is a 64.26\% improvement, which also serves as its backbone, indicating that gains arise from structured reasoning and contextual modeling rather than backbone capacity alone. \mm{Expert assessment and external dataset evaluations further demonstrate robustness under domain shift, while highlighting challenges in modeling implicit clinical nuance. Overall, \model\ provides a clinically grounded framework for evaluating therapeutic fidelity in AI mental health systems.} 

\end{abstract}

\begin{figure}[t]
    \centering
    \includegraphics[width=\columnwidth]{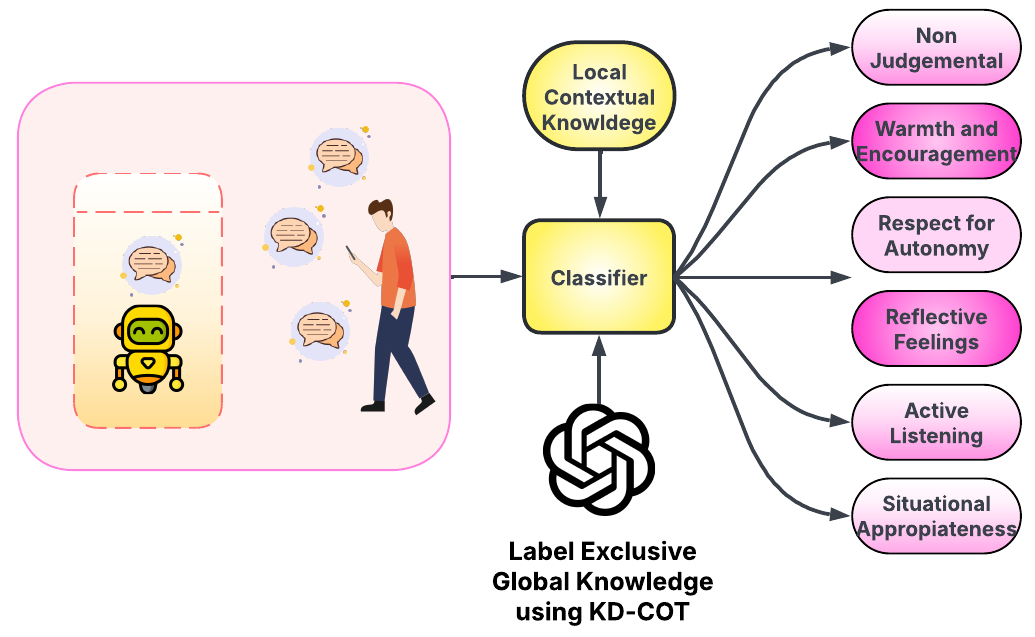}
    \caption{Our problem statement involves the ordinal classification of mental health conversation focusing on six therapeutic principles. We employ retrieval context generation along with knowledge distillation using a chain-of-thought in our proposed method, \model, ordinal classification of therapist response.}
    \label{fig:example}
    \vspace{-2.5mm}
\end{figure}

\section{Introduction}
The integration of Artificial Intelligence (AI) in mental health support has gained considerable traction, driven by its potential to deliver accessible~\cite{babu2024artificial}, scalable~\cite{almakinah2024enhancing}, and cost-effective care~\cite{cross2024use}. Conversational systems ranging from simple rule-based bots to large language models have been deployed in numerous support applications~\cite{song2024typingcureexperienceslarge}. Recent studies show that $41\%$ of support seekers have turned to conversational agents, with over $80\%$ relying on LLMs such as ChatGPT rather than clinically validated therapeutic tools\footnote{\url{https://thehemingwayreport.beehiiv.com/p/research-report-mental-health-in-an-ai-world}}. Prior work shows that lay participants often rate ChatGPT-generated therapeutic responses as comparable to or even more favorable than those written by trained clinicians~\cite{hatch2025eliza}. This suggests that standard human evaluations may be driven by surface-level qualities such as fluency or positivity, rather than clinically grounded appropriateness, thereby highlighting limitations in current evaluation practices~\cite{chang2025red}.

Although LLMs demonstrate strong fluency and contextual coherence, their application in mental health raises concerns about clinical alignment. Effective therapy depends not only on linguistic quality but also on adherence to core therapeutic principles, such as non-judgmental acceptance, reflective listening, emotional validation, and respect for client autonomy, which are essential for trust and safety. Despite their sophistication, LLM-generated responses often fall short of professional standards. \textit{Therapeutic principles}~\cite{sutton2017learning} formalize these guidelines for appropriate, context-sensitive counseling.

Recent works have explored empathetic and emotionally aware agents~\cite{Goel_2021, zaranis2021empbott5basedempatheticchatbot}, but do not explicitly address therapeutic alignment. Most approaches rely on surface-level cues or rule-based templates, which fail to capture the structured, principle-driven engagement. Critically, little work evaluates conversational models against core therapeutic principles using grounded, interpretable frameworks.

To support structured evaluation of therapeutic alignment, we build on the publicly available HOPE dataset~\cite{malhotra2022speaker}, which contains therapist client conversations across diverse counseling scenarios. We extend HOPE by introducing fine-grained expert annotations for therapeutic alignment, resulting in \dataset, a benchmark comprising 10,172 therapist utterances across 167 sessions. Each utterance is annotated with \emph{six independent labels}, one for each core therapeutic dimension, using a 5-point ordinal scale ranging from {\em strong negative} ($-2$) to {\em strong positive} ($+2$). The six dimensions, derived from established counseling theory~\cite{sutton2017learning}, include: (a) \textit{non-judgmental acceptance}, (b) \textit{warmth and encouragement}, (c) \textit{respect for autonomy}, (d) \textit{active listening}, (e) \textit{reflective understanding}, and (f) \textit{situational appropriateness}.

We further propose \model\footnote{\model: {\bf C}linical {\bf A}lignment and {\bf R}easoning {\bf E}valuation}, a clinically grounded evaluation framework that emulates expert reasoning by integrating conversational context and exemplar-driven therapeutic knowledge. \model\ encodes conversational history through a {\em relevant context} module and leverages exemplar interactions and structured reasoning via a {\em knowledge distillation}-cum-{\em chain-of-thought} (KD-CoT) module. Our evaluation across four categories of baselines demonstrates the effectiveness of \model, achieving a weighted F1-score of 63.34 and outperforming the strongest baseline (Qwen3, also used as the backbone within CARE) by +64.26\%. \mm{We further support our findings with extensive quantitative, qualitative, and expert analyses, including evaluations on external expert-guided datasets to assess generalization.} Figure~\ref{fig:example} illustrates a high-level overview of our approach.

\paragraph{\bf Problem Formulation.} 
We formulate the task as a multi-label ordinal classification problem over therapist utterances in a patient–therapist conversation $\mathcal{C} = [p_1, u_1, p_2, u_2, \cdots, p_n, u_n]$, where $p_t$ and $u_t$ denote the patient’s and therapist’s utterances, respectively. Each therapist utterance $u_t$ is evaluated along six therapeutic dimensions, with labels $y_i \in \{-2, -1, 0, +1, +2\}$ reflecting ordinal alignment. Given a local context window, the model predicts these labels while optimizing a loss that jointly captures ordinal structure and classification fidelity. Our key contributions are as follows:

\begin{itemize}[leftmargin=*, noitemsep]
\item We introduce a novel task for evaluating therapist responses against six core therapeutic principles.
\item We release \textbf{\dataset}, an expert-guided benchmark with fine-grained ordinal labels for principled evaluation of clinical dialogue.
\item We propose \textbf{\model}, a clinically grounded evaluation framework that integrates local conversational context and exemplar-based reasoning for therapeutic assessment, and \mm{demonstrate that its advantages persist under cross-dataset evaluation on trauma-focused \cite{suhas2025thousand} and CBT based \footnote{https://cheeseburgertherapy.org/} benchmarks.}
\end{itemize}

\paragraph{Reproducibility.} Code and dataset is open-sourced on \href{https://github.com/flamenlp/FAITH-M-CARE}{
  \faGithub\ GitHub \faExternalLink*\ }.

\section{Related Work}
Mental health has emerged as a critical global concern, motivating interdisciplinary research on AI-powered diagnostic tools and conversational support systems~\cite{cruz2025artificial}. While significant progress has been made in detection, assessment, and automated support delivery, comparatively little work examines whether AI-generated responses adhere to established therapeutic standards an omission that is particularly consequential in emotionally sensitive settings.

\begin{table*}[t]
\centering
\resizebox{\textwidth}{!}{
\begin{tabular}{l p{50em}}
\hline
\# & \textbf{Therapeutic Principles}\\
\hline

1 & {\bf Non-Judgmental Acceptance}: This principle refers to the therapist’s use of language that accepts the patient’s emotions and experiences without criticism, blame, or dismissal. \underline{\em Pos Indicator}: The therapist uses accepting language, demonstrating empathy toward the patient’s feelings, while \underline{\em Neg Indicator}: the therapist uses dismissive language.\\

& \hspace*{1em}$\rightarrow$ {\color{blue}\textbf{Pos Example}}:  
{\em ``It’s okay to feel that way \textbf{[SP]}". "You have the right to your feelings".}
 \\

& \hspace*{1em}$\rightarrow$ {\color{red}\textbf{Neg Example}}:  
{\em "That’s not a healthy way to feel \textbf{[SN]}". "You shouldn’t think like that \textbf{[SN]}".}\\ \hline

2 & {\bf Warmth \& Encouragement}: This principle captures the therapist’s expression of care, support, and Pos regard toward the patient. \underline{\em Pos Indicator}: The therapist uses encouraging language to convey support and empathym while \underline{\em Neg Indicator}: the therapist uses emotionally distant language.\\

& \hspace*{1em}$\rightarrow$ {\color{blue}\textbf{Pos Example}}:  
{\em "I’m glad you shared that \textbf{[MP]}". "You’re doing great by opening up. \textbf{[SP]}".}
 \\

& \hspace*{1em}$\rightarrow$ {\color{red}\textbf{Neg Example}}:  
{\em "I don’t think discussing this will help you right now."\textbf{[SN]}".}\\ \hline

3 & {\bf Respect for Autonomy}: This principle reflects the therapist’s acknowledgment of the patient’s right to make independent decisions. \underline{\em Pos Indicator}: The therapist uses language that supports the patient’s choices, while \underline{\em Neg Indicator}: the therapist uses directive language that overrides the patient’s preferences.\\

& \hspace*{1em}$\rightarrow$ {\color{blue}\textbf{Pos Example}}:  
{\em "What do you think would be best for you? \textbf{[SP]}".}
 \\

& \hspace*{1em}$\rightarrow$ {\color{red}\textbf{Neg Example}}:  
{\em "You need to do this. \textbf{[SN]}". "This is the only solution. \textbf{[SN]}".}\\ \hline

4 & {\bf Active Listening}: This principle reflects the therapist’s attentive engagement with the patient’s verbal and emotional cues. \underline{\em Pos Indicator}: The therapist uses reflective responses that acknowledge the patient’s statements, while \underline{\em Neg Indicator}: the therapist uses inattentive responses that ignore or redirect the patient’s concerns.\\

& \hspace*{1em}$\rightarrow$ {\color{blue}\textbf{Pos Example}}:  
{\em "It sounds like you’re feeling frustrated. \textbf{[SP]}".}
 \\

& \hspace*{1em}$\rightarrow$ {\color{red}\textbf{Neg Example}}:  
{\em "Let’s focus on something else. \textbf{[MN]}".}\\ \hline

5 & {\bf Reflective Feelings}: This principle captures the therapist’s ability to identify and reflect the patient’s emotional state. \underline{\em Pos Indicator}: The therapist uses language that accurately acknowledges the patient’s emotions, while \underline{\em Neg Indicator}: the therapist uses responses that ignore or misinterpret the patient’s feelings.\\

& \hspace*{1em}$\rightarrow$ {\color{blue}\textbf{Pos Example}}:  
{\em "You seem really upset about that experience. \textbf{[SP]}".}
 \\

& \hspace*{1em}$\rightarrow$ {\color{red}\textbf{Neg Example}}:  
{\em "So you quit your job last week. \textbf{[MN]}". "What’s your plan now? \textbf{[MN]}".}\\ \hline

6 & {\bf Situational Appropriateness}: This principle captures the therapist’s ability to respond in alignment with the conversational context and the patient’s emotional needs. \underline{\em Pos Indicator}: The therapist uses context-aware responses aligned with the situation, while \underline{\em Neg Indicator}: the therapist uses responses that are irrelevant or disconnected from the context.\\

& \hspace*{1em}$\rightarrow$ {\color{blue}\textbf{Pos Example}}:  
{\em "Fabulous. Okay. So would you mind if we spend a few minutes going through that  now?\textbf{[MP]}".}
 \\

& \hspace*{1em}$\rightarrow$ {\color{red}\textbf{Neg Example}}:  
{\em "Okay. Did you have any other questions about what you read on that shape?
\textbf{[MN]}".}\\ \hline

\end{tabular}}
\caption{Definitions of the six therapeutic principles used in \dataset, accompanied by representative Pos and Neg therapist utterances. All examples are annotated using a five-point ordinal scale, ranging from Strong Neg to Strong Pos, with Mild Neg and Mild Pos indicating intermediate levels of therapeutic alignment.}
\label{tab:faithfulness_examples}
\end{table*}


\paragraph{\bf AI Applications in Mental Health.}
A substantial body of work focuses on using AI to detect mental health issues from textual data. Studies have used social media and clinical transcripts to identify depressive or anxiety-related symptoms \cite{10.1145/3696410.3714778} using supervised learning and lexicon-based techniques \cite{yadav2020, cha2022lexicon, david}. The DAIC corpus was also introduced, incorporating verbal and non-verbal signals to support multimodal depression detection. While impactful for early screening, these systems are not designed for interactive therapeutic use and often lack contextual reasoning or dialogic modeling \cite{gratch2014daic}.

\paragraph{\bf Conversational Systems in Mental Health.}
Conversational agents are increasingly explored for supportive roles in mental health contexts \cite{srivastava2025trustmodelingcounselingconversations}. The coherence of LLM responses in therapeutic dialogues has been examined~\cite{song2024typingcureexperienceslarge}, and ChatGPT responses have been evaluated against therapeutic principles~\cite{hatch2025eliza}. Other efforts focus on counseling summarization~\cite{srivastava, chen2006query, srivastava2024knowledge}, dialogue understanding~\cite{malhotra2022speaker, saha2021emotion}, and response generation~\cite{srivastava2023response, lippe2020diversifying}. End-to-end systems such as CareBot~\cite{crasto2021carebot} aim to offer structured support. However, most evaluations emphasize fluency, empathy, or informativeness, relying on general-purpose metrics or subjective judgments rather than clinically grounded assessment.

\paragraph{\bf Empathy in Mental Health Interactions.}
Empathy is widely recognized as central to therapeutic communication. Prior work has explored computational empathy in online forums~\cite{sharma2021towards}, peer-support conversations~\cite{srivastava2025critical}, and dialogue systems~\cite{cai2024empcrl, srivastava2025sentimentguidedcommonsenseawareresponsegeneration}. Approaches range from affective classification to empathy-driven generation using transformer models~\cite{Goel_2021, zaranis2021empbott5basedempatheticchatbot}. However, empathy alone does not guarantee therapeutic safety or professional adherence, as it may overlook essential principles such as respect for autonomy or situational appropriateness.

\section{Dataset: \dataset}
In this section, we discuss the dataset construction pipeline. \dataset\ is derived from the publicly available HOPE dataset~\cite{malhotra2022speaker}, with each therapist utterance annotated using six ordinal labels based on therapeutic principles from {\em Learning to Counsel}~\cite{sutton2017learning}. To ensure reliability and clinical validity, expert clinicians developed the annotation guidelines and validated labels at multiple stages. The following subsections detail the dataset construction process.

\subsection{Annotation Guidelines}
We adopt a structured annotation process to evaluate therapist responses in \dataset\ across six core therapeutic principles. In collaboration with domain experts, we develop annotation guidelines to reflect clinically grounded counseling practices. We assign an ordinal score for each therapist's utterance, indicating the degree of alignment with a given therapeutic principle, ranging from \textit{strong negative} ($-2$) to \textit{strong positive} ($+2$). A score of ($0$) denotes a \textit{neutral} case in which the principle is neither expressed nor contextually required. Positive scores ($+1$, $+2$) indicate a mild to strong presence of the principle, while negative scores ($-1$, $-2$) are assigned when the utterance contradicts an expected therapeutic principle. 
We define the undertaken therapeutic principles as follows and list a few examples in Table \ref {tab:faithfulness_examples}.

\subsection{Annotation Process}
We conduct annotation in a structured, phase-wise manner to ensure reliability and consistency. Three annotators participated in the process, including two primary annotators and one senior supervising annotator. In addition, an expert clinical psychologist provided domain-specific guidance and validated the annotations.

\paragraph{Phase-1 (Training).} Annotators were trained over approximately three weeks. They were first introduced to the annotation guidelines and then participated in multiple rounds of annotation and discussion on a subset of 100 therapeutic conversations. During these sessions, the senior moderator and domain expert resolved disagreements and clarified ambiguities. Cohen’s kappa score \cite{cohen1960coefficient} was computed after each round, and training concluded upon achieving an overall kappa of $0.69$ across all therapeutic dimensions (\textit{NJ}: $0.76$, \textit{W\&E}: $0.70$, \textit{RF}: $0.64$, \textit{AL}: $0.65$, \textit{RA}: $0.73$, and \textit{SA}: $0.70$).

\paragraph{Phase-2 (Annotation).} Annotators labeled the full dataset over a period of 3 months. Their progress was periodically reviewed by peer annotators to ensure consistency with the guidelines. Final labels were consolidated by the experts, who resolved any tie-breaks between annotators.

\paragraph{Phase-3 (Validation).} Following the annotation, the expert counselor validated the labels to ensure clinical accuracy and consistency with therapeutic principles and expert-level interpretations.

\subsection{Data Analysis}
\dataset\ comprises 10,172 utterances from 167 dialogues and is annotated with six ordinal labels corresponding to core therapeutic principles. The dataset is split session-wise into \textit{train} (6,902 utterances from 116 sessions), \textit{validation} (949 utterances from 17 sessions), and \textit{test} (2,321 utterances from 34 sessions) sets in a \textit{70:10:20} ratio.
The detailed distribution for each therapeutic dimension is presented in Table~\ref{tab:data_stats}. 
\begin{table}[H]
\centering
\scriptsize  
\setlength{\tabcolsep}{6pt} 
\begin{tabular}{lccccc}
\toprule
\textbf{Therapeutic Dimension} & \textbf{SN} & \textbf{MN} & \textbf{Neu} & \textbf{MP} & \textbf{SP} \\
\midrule
Non-Judgmental Acceptance   & 107 & 413 & 624 & 1133 & 771 \\
Warmth and Encouragement    & 42 & 217 & 733 & 1134 & 921 \\
Respect for Autonomy        & 112 & 208 & 502 & 1401 & 825 \\
Active Listening            & 54 & 89 & 546 & 1458 & 900 \\
Reflective Feelings         & 84 & 155 & 116 & 1056 & 589 \\
Situational Appropriateness & 79 & 151 & 290 & 1279 & 1248 \\
\bottomrule
\end{tabular}
\caption{Label-wise distribution of utterances in \dataset. Labels range from \textit{strong negative} (SN: -2) to \textit{neutral} (neu: 0) to \textit{strong positive} (SP: +2). }
\label{tab:data_stats}
\vspace{-4mm}
\end{table}

\begin{figure}[t]  
    \centering
    \includegraphics[width=\columnwidth]{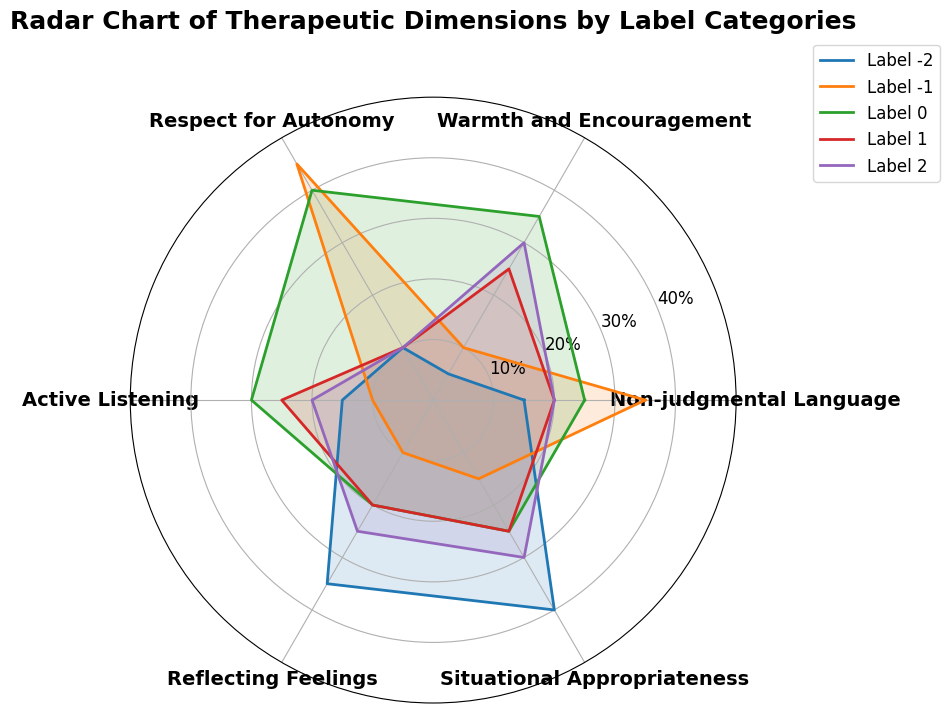}
    \caption{Distribution of ordinal labels across therapeutic dimensions in the \dataset\ dataset.}
    \label{fig:dataset_distribution}
    \vspace{-4mm}
\end{figure}

\paragraph{\bf Therapeutic Adherence.} The distribution of ordinal labels across therapeutic dimensions is shown in Figure~\ref{fig:dataset_distribution}. It reflects the strong adherence of therapists to therapeutic principles. The dataset reveals that negative labels (SN and MN) are rare, while the majority of responses are neutral or positive, indicating consistent alignment with counseling best practices. This highlights the dataset's reliability in evaluating therapist-client interactions.

\begin{figure*}[t]
    \centering
    \includegraphics[width=0.9\textwidth]{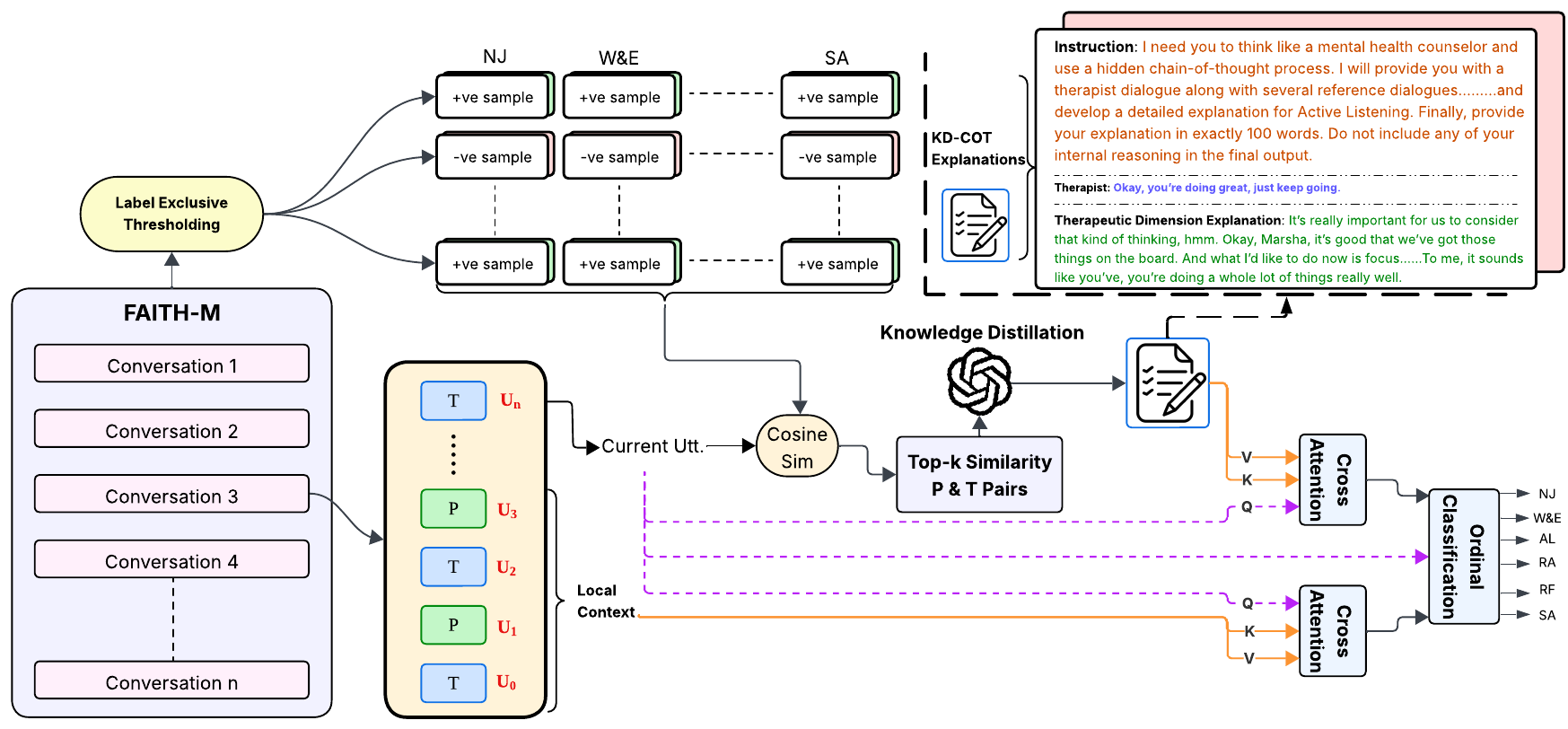}
    
    \caption{Proposed framework \model: Therapist utterances are contextualized using intra-conversational self-attention, followed by expert-informed reasoning through knowledge-distilled chain-of-thought explanations. A fusion module integrates contextual, semantic, and distilled knowledge representations, enabling ordinal classification across six core therapeutic principles.}
    \label{fig:arch}
    \vspace{-2mm}
\end{figure*}

\section{Methodology}
Therapeutic assessment depends on conversational context, as utterance meaning is shaped by prior dialogue. Accordingly, \model\ encodes local conversation history through a {\em relevant context module} and incorporates clinical reasoning via {\em chain-of-thought–based knowledge distillation}. Figure~\ref{fig:arch} illustrates the overall architecture.

\subsection{Relevant Context Module}
We construct a local context window \(\mathbf{C}\) for each therapist utterance \(u_t\), containing the utterance itself and \(k\) preceding conversational turns: \{$p_{t-k}, u_{t-k}, \ldots, p_{t},  u_{t}\}$, where \( p_i \) and \( u_i \) denote the patient's and therapist's utterance at turn \( i \), respectively. This window provides immediate conversational context, enabling the encoder’s self-attention~\cite{vaswani2017attention} to capture dependencies between prior therapist interventions and the patient’s evolving state.
The resulting representation \(\mathbf{R}_{\text{ctx}}\) serves as one of the input streams to the final prediction layer, capturing the situational grounding necessary for accurate therapeutic alignment assessment.

\subsection{KD-CoT}

To explicitly embed clinical reasoning and therapeutic knowledge into our model, we utilize a knowledge distillation approach based on Chain-of-Thought prompting. This module enables the model to learn not only from raw examples but also from structured rationales that emulate expert reflection. The KD-CoT pipeline comprises the following phases:

\paragraph{Label-Exclusive Exemplar Sets.}
To distinguish positive and negative therapeutic alignments for each principle, we construct label-exclusive exemplar sets from the \emph{training split} of \dataset. We retain only therapist utterances labeled as strongly positive or strongly negative, excluding neutral and borderline cases to reduce ambiguity and capture canonical patterns of adherence and violation. For each principle, we curate two exemplar pools of patient–therapist pairs, where the therapist response immediately follows the patient utterance to ensure contextual coherence. Utterances from the validation and test splits are strictly excluded from exemplar construction and are used only for model selection and final evaluation.

\paragraph{Reference Retrieval.} We embed each patient-therapist pair using a Sentence Transformer \cite{reimers2019sentence}. For a given test instance, we compute cosine similarity between it's embedding and those of the reference sets. We then retrieve the top two most semantically similar reference pairs per dimension and polarity. The aim is to keep the selected examples both contextually relevant and therapeutically aligned. The retrieved exemplars from the previous step are passed to GPT-4o along with a structured prompt\footnote{Refer to Appendix to view prompts (cf Figure~\ref{fig:explanation_prompt}).}. GPT-4o is used to generate dimension-specific CoT explanations grounded in therapeutic norms Figure~\ref{fig:few_zero_prompts} (c.f. Appendix).

\paragraph{Knowledge Encoding.} 
The retrieved exemplar pairs are provided to a LLM via structured, few-shot prompts to elicit dimension-specific explanations grounded in therapeutic norms. For each therapeutic principle, the model generates concise rationales explaining why a therapist’s response demonstrates alignment or constitutes a violation. These explanations act as expert-like reasoning traces that capture clinically salient cues beyond surface lexical patterns. In our primary setup, we use GPT-4o as the reasoning teacher for this knowledge distillation step and report its results in the main experiments. The generated rationales are concatenated with the current therapist utterance and encoded using a Qwen3 \cite{yang2025qwen3} model, producing a distilled knowledge representation $\mathbf{R}_{\text{KD}}$, which is integrated with other model streams for final ordinal prediction.

\subsection{Ordinal Classification Block}

We integrate the context-aware representation $R_{\text{ctx}}$, the distilled knowledge representation $R_{\text{KD}}$, and the utterance embedding $r_t$ using a cross-attention-based fusion mechanism. Specifically, $r_t$ is used as the query, while $R_{\text{ctx}}$ and $R_{\text{KD}}$ serve as key-value pairs, enabling the model to attend over contextual and knowledge signals. We then pass the resulting fused representation to an ordinal classification head to predict alignment scores across five levels. To jointly model ordinal structure and categorical accuracy, we use a hybrid loss:
\begin{equation}\label{eq2}
\mathcal{L} = \alpha \cdot \text{MSE}(\hat{y}, y) + \beta \cdot \text{CE}(\hat{y}, y),
\end{equation}
where MSE captures ordinal distance and CE models class-wise accuracy. We tune $\alpha, \beta \in {0.3, 0.5, 0.7}$ on the validation set and observe stable performance, with $\alpha = \beta = 0.5$ performing best. We use this setting in all experiments to balance ordinal consistency and classification accuracy.

\section{Experiments and Results}
Here, we compare the \model performance against baseline methods and present a detailed ablation study to analyze the contribution of individual components. We further provide quantitative and qualitative analyses, including error analysis and human expert evaluations, to assess performance and alignment with core therapeutic principles.

\paragraph{\bf Baselines.}
We benchmark \model\ against a diverse set of 15 baselines, comprising 4 prompt-based and 11 trainable models, evaluated under a unified experimental setup. The baseline methods can be viewed in Table \ref{tab:results}, and we include more details pertaining to the baseline checkpoints in Appendix \ref{app:baseline}. All baselines use the same local context window ($k=2$) and, where applicable, are trained with the same ordinal-aware loss function (Equation~\ref{eq2}).

\begin{table}[t]
    \centering
    \renewcommand{\arraystretch}{1.0}
    \setlength{\tabcolsep}{4pt} 

\resizebox{\columnwidth}{!}{
\begin{tabular}{llcccc}
    \toprule
    & \textbf{Models} & \textbf{Acc} & \textbf{Pre} & \textbf{Re} & \textbf{F1$_w$} \\
    \midrule

      & Mental Llama   & 27.94  & 8.69  & 27.74  & 12.97  \\
     \multirow{-2}{*}{\rotatebox{90}{\textbf{ZS}}}  & GPT-4o          & 31.09  & 36.19  & 31.09  & 30.49  \\ \midrule

    & Mental Llama   & 27.74  & 8.62  & 27.74  & 12.97  \\
 \multirow{-2}{*}{\rotatebox{90}{\textbf{FS}}} & GPT-4o
 & 30.44  & 32.28  & 30.44  & 22.40  \\
\midrule
    \multirow{4}{*}{\rotatebox{90}{\textbf{Enc}}} & MentalBERT    & 23.85  & 30.05  & 23.85  & 25.73  \\
    & RoBERTa       & 30.09  & 33.08  & 30.09  & 30.72  \\
    & ALBERT        & 42.32  & 41.58  & 42.32  & 34.49  \\
    & DeBERTa       & 33.79  & 35.32  & 33.79  & 34.52  \\
    
    \midrule
    & MentalBART   & 41.26  & 46.08  & 41.27  & 27.30  \\
 \multirow{-2}{*}{\rotatebox{90}{\textbf{En-De}}} & BART          & 33.34  & 34.61  & 33.34  & 33.96  \\
    
    \midrule
    \multirow{4}{*}{\rotatebox{90}{\textbf{Dec}}} & Gemma     & 36.56 & 37.23  & 36.56  & 36.88     \\ 
    & Phi4        &   36.89   &   37.91   & 35.81     & 37.37     \\
    
    & LLaMA 3.1    & 43.36  & 42.81  & 43.46  & 37.73  \\
     \rowcolor{gray!15} & {LLaMA 3.2} & {44.91} & {44.78} & {44.91} & {37.90} \\
     \rowcolor{gray!15} & {CARE-LLaMA 3.2} & {62.07} & {64.11} & {62.07} & {63.07} \\
    
    \midrule
    \rowcolor{green!20} & \textbf{\mm{Qwen3}} & \textbf{\mm{45.47}} & \textbf{\mm{45.10}} & \textbf{\mm{45.38}} & \textbf{\mm{38.56}} \\
    
    \rowcolor{green!20} & \textbf{\mm{\model-Qwen3}} & \textbf{\mm{63.30}} & \textbf{\mm{64.05}} & \textbf{\mm{62.65}} & \textbf{\mm{63.34}} \\
\midrule
& $\Delta_{Baseline}(\%)$ & \textcolor{blue}{$\uparrow 39.21\%$} & \textcolor{blue}{$\uparrow 42.03\%$} & \textcolor{blue}{$\uparrow 38.05\%$} & \textcolor{blue}{$\uparrow 64.26\%$} \\

    \midrule
& \model-Qwen3\ -- (LC) & 57.08 & 58.06 & 57.02 & 57.2 \\
& \model-Qwen3\ -- (LE) & 53.81 & 54.04 & 53.02 & 53.08 \\
    \bottomrule
\end{tabular}
}
    \caption{Comparative analysis across multiple model classes.
LLaMA-3.2 and Qwen-3 represent the strongest-performing baseline backbones, and \model\  yields consistent and substantial improvements over both, indicating that gains are not specific to a single base model. LC: label-context; LE: label-exclusive; ZS: zero-shot; FS: few-shot; Enc: encoder; Dec: decoder.}
    \label{tab:results}
\end{table}

\begin{table}[t]
\centering
\resizebox{\columnwidth}{!}{
\begin{tabular}{llccccccc}
\toprule
\textbf{\#} & \textbf{Param} & \textbf{Acc.} & \multicolumn{3}{c}{\textbf{Macro}} & \multicolumn{3}{c}{\textbf{Weighted}}\\
\cmidrule(lr){4-6} \cmidrule(lr){7-9}
 & & & \textbf{F1} & \textbf{Prec.} & \textbf{Rec.} & \textbf{F1} & \textbf{Prec.} & \textbf{Rec.} \\
\midrule
\multirow{5}{*}{\rotatebox{90}{\textbf{GPT-4o}}} 
& $k = 1$ & 55.66 & 53.77 & 54.68 & 52.89 & 56.03 & 56.42 & 55.65 \\
& $k = 2$ & 55.35 & 53.26 & 53.56 & 52.98 & 56.04 & 56.79 & 55.32 \\
& \cellcolor{gray!20}$k = 3^\Theta$ & \cellcolor{gray!20}\textbf{56.37} & \cellcolor{gray!20}\textbf{55.18} & \cellcolor{gray!20}\textbf{56.27} & \cellcolor{gray!20}\textbf{54.15} & \cellcolor{gray!20}\textbf{57.55} & \cellcolor{gray!20}\textbf{58.84} & \cellcolor{gray!20}\textbf{56.32} \\
& $k = 4$ & 54.99 & 51.38 & 52.15 & 50.64 & 55.04 & 55.84 & 54.97 \\
& $k = 5$ & 48.86 & 43.41 & 44.12 & 42.74 & 49.20 & 49.56 & 48.85 \\
\midrule
\multirow{5}{*}{\rotatebox{90}{\textbf{Mental Llama}}} 
& $k = 1$ & 45.36 & 43.25 & 44.51 & 42.07 & 46.56 & 47.85 & 45.34 \\
& $k = 2$ & 45.01 & 43.30 & 44.68 & 42.01 & 46.36 & 47.74 & 44.97 \\
& $k = 3$ & \textbf{45.97} & \textbf{43.76} & \textbf{45.19} & \textbf{42.32} & \textbf{47.21} & \textbf{48.64} & \textbf{45.88} \\
& $k = 4$ & 43.67 & 43.11 & 44.57 & 41.76 & 45.18 & 46.87 & 43.64\\
& $k = 5$ & 43.18 & 40.92 & 41.23 & 40.63 & 44.78 & 46.54 &  43.15\\
\bottomrule
\end{tabular}
}
\caption{Ablation Study for relevant context ({\em k}) and KD-COT Modules (with GPT-4o and MentalLlama) on combined positive ({\em +ve}) and negative ({\em -ve}) indicators. Results with more indicators and local context are included in Appendix (c.f. Table \ref{tab:ablation_study}).}
\label{tab:ablation_pos_neg}
\vspace{-3mm}
\end{table}

\paragraph{\bf Performance Comparison.}
Table~\ref{tab:results} reports performance on the FAITH-M benchmark. Encoder and encoder–decoder models capture coarse therapeutic signals but struggle with fine-grained ordinal distinctions, particularly in mid-spectrum categories. Decoder-only LLMs perform better, but their predictions often rely on surface-level empathy cues, limiting robustness. \mm{\model\ achieves substantial gains over the strongest decoder-only baseline, Qwen3, which also serves as the backbone of the CARE architecture. These improvements are most pronounced in weighted F1, reflecting better handling of both frequent and infrequent ordinal labels. Both Qwen3 and LLaMA~3.2 are included due to their competitive baseline performance, highlighting that CARE consistently improves across different LLM backbones. To enable a fairer comparison, we also fine-tune MentalLLaMA using the same KD-CoT supervision employed within CARE, resulting in a marked improvement over its zero-shot counterpart.} Label-wise performance is reported in (Appendix Table \ref{tab:labelWiseResults}, with the strongest results observed for Non-Judgmental Language and comparatively lower scores for Warmth and Encouragement and Reflecting Feelings. Overall, \model\ continues to outperform all baselines, indicating that CARE’s gains arise from the joint integration of contrastive exemplar-based reasoning and explicit local contextual encoding.

\paragraph{\bf Ablation Study.}
We conduct an ablation study to assess the contribution of core CARE components, including local contextual encoding and the KD-CoT module. Here, GPT-4o and MentalLLaMA are used as reasoning teachers to generate Chain-of-Thought rationales within the KD-CoT framework, while the final predictions are produced by the student model. Table~\ref{tab:ablation_pos_neg} reports performance as we progressively add contextual grounding, label-exclusive exemplar retrieval, and contrastive KD-CoT reasoning, showing consistent improvements with each component. The largest gains arise from combining context with polarity-aware exemplar reasoning, indicating that CARE’s effectiveness stems from structured reasoning rather than dialogue context or prompt supervision alone. Figure~\ref{fig:kd_cot_ablation} visualizes these trends, where moderate context sizes ($k=2$–$3$) yield peak performance and larger windows show diminishing returns. Additional ablation analyses and implementation details are provided in (Appendix ~\ref{sec:additionalablation}).

\begin{table}[t]
\centering
\resizebox{\columnwidth}{!}{
\begin{tabular}{lccccccc} 
\toprule
\multirow{2}{*}{\textbf{Model}} & \textbf{Acc.} & \multicolumn{3}{c}{\textbf{Macro}} & \multicolumn{3}{c}{\textbf{Weighted}}\\
\cmidrule(lr){3-5} \cmidrule(lr){6-8}
 & & \textbf{F1} & \textbf{Prec.} & \textbf{Rec.} & \textbf{F1} & \textbf{Prec.} & \textbf{Rec.} \\
\midrule
\multicolumn{8}{l}{\textit{\textbf{Dataset: PTSD}}} \\ 
\midrule
GPT-4o & 38.26 & 25.53 & 28.79 & 28.49 & 35.06 & 37.09 & 38.26 \\
Qwen & 38.00 & 16.80 & 19.10 & 25.30 & 31.10 & 34.90 & 38.00 \\
Llama 3.2 & 29.80 & 8.80 & 8.60 & 18.90 & 17.50 & 14.50 & 29.80 \\
\rowcolor{gray!20} \textbf{CARE} & \textbf{52.34} & \textbf{46.80} & \textbf{48.80} & \textbf{46.50} & \textbf{50.90} & \textbf{51.30} & \textbf{52.40} \\
\midrule
\multicolumn{8}{l}{\textit{\textbf{Dataset: CheeseBurger}}} \\ 
\midrule
GPT-4o & 26.33 & 12.62 & 19.61 & 10.14 & 31.50 & 42.22 & 26.33 \\
Qwen & 28.05 & 11.25 & 10.71 & 16.58 & 26.94 & 29.35 & 28.05 \\
Llama 3.2 & 23.15 & 9.01 & 16.98 & 17.89 & 18.95 & 39.13 & 23.15 \\
\rowcolor{gray!20} \textbf{CARE} & \textbf{48.96} & \textbf{20.46} & \textbf{28.28} & \textbf{20.07} & \textbf{50.30} & \textbf{65.25} & \textbf{48.96} \\
\bottomrule
\end{tabular}
}
\caption{\mm{Zero-shot performance comparison. The CARE method consistently outperforms baselines on both PTSD and CheeseBurger datasets.}}
\label{tab:generalizablity}
\end{table}
\section{Analysis}
Here, we discuss analysis, expert evaluation, and generalizability.

\paragraph{\bf Generalizability.}
In addition to the \model's benchmarking on \dataset\ (c.f. Table \ref{tab:results}), we extended the \model's inference on two cross-dataset evaluations: PTSD \cite{suhas2025thousand} and Cheeseburger\footnote{https://cheeseburgertherapy.org/}. Table~\ref{tab:generalizablity} shows detailed results and highlights that the advantage persists under cross-dataset evaluation. On the PTSD dataset, which contains trauma-focused Prolonged Exposure therapy dialogues, and the CBT-based Cheeseburger dataset, \model\ consistently outperforms prompt-based and fine-tuned LLM baselines. Although absolute performance decreases relative to in-domain evaluation, as expected with considerable domain-shift, the relative improvements remain stable, suggesting that CARE generalizes beyond the distribution on which it was developed.

\begin{figure}[t]
    \centering
    \includegraphics[width=0.90\columnwidth]{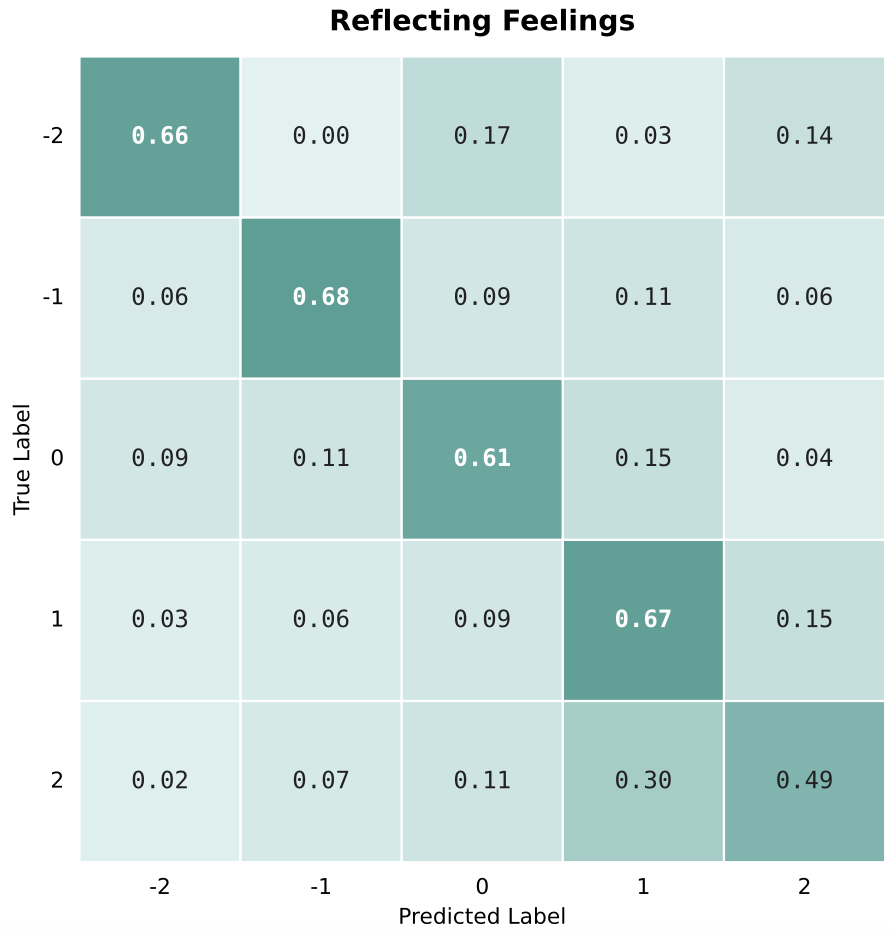}
    \caption{Confusion matrix for \textit{Reflecting Feelings}. Refer to Figure \ref{fig:confusionmatrix} for other dimensions' matrices.}
    \label{fig:rf_cf}
\end{figure}

\begin{table*}[t]
\centering
\resizebox{\textwidth}{!}{
\begin{tabular}{@{}l p{11cm}cccccc@{}}
\toprule
\textbf{Type} & \textbf{Utterance} & \textbf{NJL} & \textbf{Warmth} & \textbf{Autonomy} & \textbf{Listening} & \textbf{Reflect} & \textbf{SA} \\
\midrule
P & Well, I was raised Catholic. & & & & & & \\ 
T & So you consider yourself Catholic. But you feel like there's a lot of rules. 
& {N} / {N}
& {N} / \textcolor{red}{MP}
& {MP} / {MP}
& {N} / \textcolor{red}{SP}
& {N} / \textcolor{red}{MP}
& {MP} / \textcolor{red}{SP} \\ \midrule
P & Yeah. I mean, there's a lot of do's and don'ts. Our church was very like, hell oriented. Like, if you do this, you're gonna go to hell... & & & & & & \\
T & How do you think those spiritual values influence your daily life?
& {MP} / {MP}
& {SP} / \textcolor{red}{MP}
& {N} / \textcolor{red}{SP}
& {MP} / {MP}
& {N} / {N}
& {SP} / {SP} \\
\bottomrule
\end{tabular}}
\caption{Examples of \textcolor{red}{\em mispredictions} by \model. These errors typically arise from subtle pragmatic mismatches or an overestimation of empathic alignment. {N}: Neutral, {SP}: Strong Positive, {MP}: Mild Positive, {SN}: Strong Negative, {MN}: Mild Negative.}
\label{tab:qualitative-errors}
\vspace{-4mm}
\end{table*}


\paragraph{\bf Quantitative Analysis.}
\label{sec:confusionExplanation}

We analyze the confusion matrices to understand error patterns across therapeutic dimensions using the updated model configuration in Figure~\ref{fig:rf_cf}. For the \textit{Reflecting Feelings} dimension, the model shows reliable recognition of extreme categories, with accuracies of 66\% for Strong Negative and 49\% for Strong Positive. As expected in ordinal classification, most errors occur between adjacent labels, particularly between Mild Positive and Strong Positive (30\%), reflecting the difficulty of distinguishing subtle emotional affirmation from strong empathic reflection. Similar trends are observed across the remaining therapeutic principles (Appendix, Figure~\ref{fig:confusionmatrix}). The model consistently maintains strong diagonal mass for extreme labels, achieving approximately 65–70\% accuracy for Strong Negative and Mild Negative classes across dimensions such as \textit{Non-Judgmental Language}, \textit{Active Listening}, and \textit{Situational Appropriateness}. 
In contrast, mid-spectrum categories remain challenging, with {\em neutral} labels often confused with adjacent {\em mild positive} or {\em mild negative} classes (15-20\% misclassification, depending on the dimension).
Overall, these confusion patterns indicate that the model captures ordinal structure effectively, with errors largely constrained to neighboring categories rather than cross-polarity confusions. This behavior suggests sensitivity to graded therapeutic alignment while highlighting the inherent ambiguity of mid-spectrum judgments.

\paragraph{\bf Error Analysis.}
Table~\ref{tab:qualitative-errors} presents cases where \model\ diverges from expert annotations. Errors typically arise from misinterpreting subtle therapeutic cues, such as emotionally grounded paraphrasing, which the model sometimes labels as overly positive or neutral, consistent with patterns observed in earlier analyses. Across $6$ utterances, the model deviated from gold labels in $7$ out of $18$ dimension-level predictions, usually by a margin of $\pm1$. For example, the utterance \textit{“So you consider yourself Catholic. But you feel like there’s a lot of rules.”} was annotated as {\em neutral} for \textit{non-judgmental acceptance}, while the model predicted {\em mild-positive}, reflecting a tendency to overestimate surface-level reflectiveness as affirming. Errors were most frequent for \textit{situational appropriateness} and \textit{reflective feelings} (3/6 cases each), whereas \textit{non-judgmental acceptance} and \textit{respect for autonomy} showed higher agreement (4/6 correct each). The model occasionally carries forward prior sentiment across topic shifts, leading to misalignment in \textit{situational appropriateness} and highlighting challenges in modeling mid-range ordinal labels.

\paragraph{\bf Expert Assessment.}
We consult a licensed clinical psychologist to independently review model predictions and assess alignment with human therapeutic judgments across all dimensions. The expert agreed with the model’s predictions in a substantial proportion of cases. Agreement was highest for dimensions with explicit emotional and supportive cues, including \textit{warmth and encouragement} (84.0\%) and \textit{non-judgmental acceptance} (81.6\%). In contrast, agreement was lower for more interpretive dimensions such as \textit{respect for autonomy} (70.3\%) and \textit{reflecting feelings} (66.7\%), which require deeper contextual and emotional understanding. Qualitative feedback further indicates that while \model\ reliably captures overt emotional signals, it occasionally overestimates superficially supportive responses and underestimates subtle acknowledgements. The model also shows limitations when conversations shift abruptly, affecting its handling of \textit{situational appropriateness}. Detailed agreement statistics are reported in Table~\ref{tab:expert_agreement}.
\begin{table}[H]
\centering
\resizebox{\columnwidth}{!}{
\begin{tabular}{@{}cccccc@{}}
\toprule
\textbf{NJL} & \textbf{Warmth} & \textbf{Autonomy} & \textbf{Listening} & \textbf{Reflect} & \textbf{SA} \\
\midrule
81.60\% & 84.00\% & 70.30\% & 70.10\% & 66.70\% & 69.20\% \\
\bottomrule
\end{tabular}
}
\caption{Agreement score: Expert vs \model.}
\label{tab:expert_agreement}
\vspace{-3.5mm}
\end{table}

\paragraph{Discussion}
\label{sec:misalignment_appendix}

\mm{We analyze GPT-4o under zero-shot and few-shot prompting to examine the limitations of prompt-based therapeutic evaluation. While few-shot prompting yields a modest gain in average accuracy, it leads to a substantial drop in weighted F1, indicating poorer calibration across ordinal categories. This suggests that in-context examples improve format adherence without enhancing sensitivity to ordinal severity. Per-dimension results in Table~\ref{tab:zerovsfew} show stronger performance on \textit{Warmth and Encouragement} and \textit{Active Listening}, where alignment is often conveyed through surface-level cues, and weaker performance on \textit{Non-Judgmental Language} and \textit{Respect for Autonomy}, which require reasoning about implicit judgment and relational intent, as evident under zero-shot prompting (Figure~\ref{fig:gpt_zero_cf}), GPT-4o frequently collapses negative and neutral instances into the Neutral class. Few-shot prompting (Figure~\ref{fig:gpt_few_cf}) shifts predictions toward {\em mild} and {\em strong positive} labels without recovering {\em negative} or {\em neutral} categories. Figure~\ref{fig:fewshot_full_example} (c.f. Appendix) illustrates GPT-4o’s tendency to replicate exemplar patterns rather than adapt to the target context, highlighting structural limits of prompt-only evaluation for fine-grained therapeutic assessment.}
\begin{table}[t]
\centering
\resizebox{\columnwidth}{!}{
\begin{tabular}{llccccccc}
\toprule
\textbf{\#} & \textbf{In-Context} & \textbf{Dim} & \multicolumn{3}{c}{\textbf{Macro}} & \multicolumn{3}{c}{\textbf{Weighted}} \\
\cmidrule(lr){4-6} \cmidrule(lr){7-9}
& & & \textbf{F1} & \textbf{Prec.} & \textbf{Rec.} & \textbf{F1} & \textbf{Prec.} & \textbf{Rec.} \\
\midrule

\multirow{14}{*}{\rotatebox{90}{\textbf{GPT-4o}}}
& \multirow{7}{*}{Zero Shot}
& NJ       & 23.67 & 25.81 & 25.75 & 23.67 & 25.81 & 25.75 \\
& & W\&E    & 20.44 & 33.60 & 24.55 & 20.44 & 33.60 & 24.55 \\
& & RA      & 30.99 & 31.32 & 34.73 & 30.99 & 31.32 & 34.73 \\
& & AL      & 37.20 & 42.19 & 33.53 & 37.20 & 42.19 & 33.53 \\
& & RF      & 38.94 & 45.01 & 37.42 & 38.94 & 45.01 & 37.42 \\
& & SA      & 31.68 & 39.18 & 30.53 & 31.68 & 39.18 & 30.53 \\
& & \textbf{Avg.} & \textbf{30.49} & \textbf{36.18} & \textbf{31.08} & \textbf{30.49} & \textbf{36.18} & \textbf{31.08} \\

\cmidrule(lr){2-9}
& \multirow{7}{*}{Few-Shot}
& NJ       & 13.01 & 19.14 & 25.74 & 13.01 & 19.14 & 25.74 \\
& & W\&E    & 39.22 & 44.65 & 44.91 & 39.22 & 44.65 & 44.91 \\
& & RA      & 18.52 & 45.83 & 27.24 & 18.52 & 45.83 & 27.24 \\
& & AL      & 24.40 & 38.45 & 31.13 & 24.40 & 38.45 & 31.13 \\
& & RF      & 18.27 & 27.95 & 23.65 & 18.27 & 27.95 & 23.65 \\
& & SA      & 25.31 & 33.95 & 37.42 & 25.31 & 33.95 & 37.42 \\
& & \textbf{Avg.} & \textbf{23.12} & \textbf{34.99} & \textbf{31.69} & \textbf{23.12} & \textbf{34.99} & \textbf{31.69} \\




\bottomrule
\end{tabular}
}
\caption{\mm{Per-dimension macro and weighted scores for GPT-4o under zero-shot and few-shot settings. Short labels: NJ = Non-Judgmental, W\&E = Warmth \& Encouragement, RA = Respect for Autonomy, AL = Active Listening, RF = Reflecting Feelings, SA = Situational Appropriateness.
}}
\label{tab:zerovsfew}
\vspace{-2mm}
\end{table}

\section{Conclusion}
We presented \model, a context-aware framework for evaluating therapist responses across six therapeutic principles using fine-grained ordinal judgments. By integrating local conversational context with knowledge-distilled chain-of-thought reasoning, \model\ enables more reliable assessment of therapeutic alignment beyond surface-level fluency. We also introduced \dataset, an expert-guided benchmark to support principled evaluation of clinical dialogue. Our findings highlight the limitations of prompt-only methods in ordinal calibration and nuanced therapeutic reasoning, particularly for interpretive dimensions. In contrast, \model\ provides a structured and extensible evaluation approach. Future work will explore multi-turn assessment and patient-specific and cross-cultural settings.

\section{\bf Ethics and Limitations.}
While \model\ demonstrates strong performance in utterance-level therapeutic evaluation, several limitations remain. The six principles considered capture only a subset of clinical competencies; aspects such as cultural competence, trauma-informed care, crisis assessment, and intervention timing are not modeled. This scope was intentionally limited to communicative alignment rather than holistic clinical judgment. \model\ operates on individual utterances with limited local context and does not capture long-term therapeutic dynamics such as alliance building across sessions. Moreover, therapeutic appropriateness is inherently context- and culture-dependent, and interpretations of warmth or support may vary across settings. Cultural and contextual adaptation was not a primary focus and remains an important direction for future work.

Finally, \model\ is designed solely for evaluating AI-generated therapist-like responses and does not provide clinical interventions or replace professional care. All data are anonymized, annotations were conducted under expert supervision, and any real-world use should include clinician oversight and appropriate safeguards.

\section{Acknowledgment}
The authors acknowledge the support of the Infosys Foundation through CAI at IIIT-Delhi.

\bibliography{custom}

\clearpage
\appendix
\section*{Appendix}
\label{sec:appendix}



\section{Baseline Methods}
\label{app:baseline}
We benchmark \model\ against a diverse set of prompt-based and trainable baselines under a unified experimental setup. All baselines use the same local context window ($k=2$) and, where applicable, are trained with the same ordinal-aware loss function (Equation~\ref{eq2}).

\textbf{(a) Zero-shot LLMs:} GPT-4o \cite{islam2024gpt} and MentalLLaMA \cite{mentalllama} are evaluated without task-specific fine-tuning to gauge the upper bound of out-of-the-box performance.  
\textbf{(b) Few-shot LLMs:} We further extend this by evaluating GPT-4o  under few-shot prompting \ref{tab:zerovsfew}, using carefully constructed in-context examples (Appendix~\ref{fig:few_zero_prompts}) to examine whether performance improves with light in-context supervision. A detailed analysis of both zero-shot and few-shot performance including prompt formats, per-dimension scores, and failure analysis is presented in (Discussion~\ref{sec:misalignment_appendix}).
\textbf{(c) Transformer encoders:} MentalBERT \cite{mentalbert}, ALBERT \cite{albert}, DeBERTa \cite{deberta}, and RoBERTa \cite{roberta} test how well discriminative, parameter-efficient encoders capture therapeutic nuances when fine-tuned on our data; \textbf{(d) Encoder–decoders:} BART \cite{bart} and its domain-adapted variant MentalBART \cite{mentalbart} assess the utility of sequence-to-sequence pretraining for dialogue-level classification tasks;  
\textbf{(e) Decoder-only LLMs:} Qwen3 \cite{yang2025qwen3}, LLaMA 3.1/3.2 \cite{llama}, Phi-4 \cite{phi4}, and Gemma \cite{gemmateam2024gemma2improvingopen} probe the limits of autoregressive generators when repurposed for structured ordinal prediction.

\section{Experimental Setup}

\mm{All experiments were conducted on 2 NVIDIA A100 GPUs. Models were trained for 10 epochs using the AdamW optimizer with a learning rate of $1\times10^{-5}$, weight decay of 0.01, dropout of 0.2, and gradient clipping at 1.0. Training was performed with a batch size of 16, and input sequences were truncated to the maximum length supported by each backbone. A sliding window of size 6 was used to handle longer dialogues. Parameter-efficient fine-tuning was implemented using LoRA with rank $r=16$ and scaling factor $\alpha=32$. All experiments were run with a fixed random seed of 42. A complete list of hyperparameters is provided in Table~\ref{tab:hyperparameters}. Inference time varies with the choice of the KD-CoT reasoning teacher. Using open-source models (e.g., Qwen) for rationale generation yields approximately 1 second per instance with a weighted F1 of 57.75, whereas using GPT-4o increases inference time to around 5 seconds but improves weighted F1 to 63.34. This highlights a clear trade-off between efficiency and evaluation accuracy.}

\begin{table}[t]
\centering
\begin{tabular}{ll}
\hline
\textbf{Hyperparameter}      & \textbf{Value} \\ \hline
Learning Rate (lr)           & \(1 \times 10^{-5}\) \\
Adam Beta1                 & 0.9 \\
Adam Beta2                 & 0.999 \\
Dropout                    & 0.2 \\
Weight Decay               & 0.01 \\
Batch Size                 & 16 \\
Max Sequence Length        & 4096 \\
Sliding Window Size        & 6 \\
Gradient Clipping          & 1.0 \\ 
Learning Rate Scheduler    & Constant \\
Seed                       & 42 \\ \hline
LoRA Rank (r)              & 32 \\
LoRA Alpha (\(\alpha\))    & 32 \\ \hline
\end{tabular}
\caption{Detailed hyperparameters used in the experiments.}
\label{tab:hyperparameters}
\end{table}

\begin{table}[t]
\centering
\resizebox{\columnwidth}{!}{
\begin{tabular}{llccccccc}
\toprule
\textbf{\#} & \textbf{Param} & \textbf{Acc.} & \multicolumn{3}{c}{\textbf{Macro}} & \multicolumn{3}{c}{\textbf{Weighted}}\\
\cmidrule(lr){4-6} \cmidrule(lr){7-9}
 & & & \textbf{F1} & \textbf{Prec.} & \textbf{Rec.} & \textbf{F1} & \textbf{Prec.} & \textbf{Rec.} \\
\midrule
\multirow{5}{*}{\rotatebox{90}{\textbf{Rel. Context}}} 
& $k = 1$ & 49.68 & 46.80 & 47.05 & 46.56 & 50.41 & 51.24 & 49.62 \\
&$k = 2$  & \textbf{53.12} & \textbf{47.60} & \textbf{48.34} & \textbf{46.92} & \textbf{54.02} &\textbf{55.24} & \textbf{52.87} \\
& $k = 3$ & 50.58 & 47.07 & 47.86 & 46.32 & 51.26 & 52.01 & 50.55 \\
& $k = 4$ & 48.95 & 46.21 & 46.67 & 45.77 & 50.17 & 51.52 & 48.89 \\
& $k = 5$ & 47.73 & 43.98 & 44.67 & 43.32 & 48.43 & 49.24 & 47.65 \\
\midrule
\multirow{15}{*}{\rotatebox{90}{\textbf{KD-COT Module (GPT-4o)}}} 
& \multicolumn{8}{l}{\textit{$+$ve \& $-$ve labels}} \\
& $k = 1$ & 55.66 & 53.77 & 54.68 & 52.89 & 56.03 & 56.42 & 55.65 \\
& $k = 2$ & 55.35 & 53.26 & 53.56 & 52.98 & 56.04 & 56.79 & 55.32 \\
& \cellcolor{gray!20}$k = 3^\Theta$ & \cellcolor{gray!20}\textbf{56.37} & \cellcolor{gray!20}\textbf{55.18} & \cellcolor{gray!20}\textbf{56.27} & \cellcolor{gray!20}\textbf{54.15} & \cellcolor{gray!20}\textbf{57.55} & \cellcolor{gray!20}\textbf{58.84} & \cellcolor{gray!20}\textbf{56.32} \\
& $k = 4$ & 54.99 & 51.38 & 52.15 & 50.64 & 55.04 & 55.84 & 54.97 \\
& $k = 5$ & 48.86 & 43.41 & 44.12 & 42.74 & 49.20 & 49.56 & 48.85 \\
\cmidrule(lr){2-9}
& \multicolumn{8}{l}{\textit{$+$ve Labels Only}} \\
& $k = 1$ & 46.20 & 42.70 & 42.05 & 43.39 & 48.09 & 50.17 & 46.19 \\
& $k = 2$ & 49.68 & 47.23 & 47.53 & 46.94 & 51.02 & 52.47 & 49.66 \\
& $k = 3$ & \textbf{50.22} & \textbf{47.22} & \textbf{47.81} & \textbf{46.65} & \textbf{51.73} & \textbf{53.36} & \textbf{50.20} \\
& $k = 4$ & 50.12 & 47.12 & 48.43 & 45.88 & 51.14 & 52.24 & 50.09 \\
& $k = 5$ & 48.81 & 45.66 & 46.39 & 44.97 & 49.13 & 51.77 & 48.76\\
\cmidrule(lr){2-9}
& \multicolumn{8}{l}{\textit{$-$ve Labels Only}} \\
& $k = 1$ & 45.23 & 41.58 & 40.89 & 42.31 & 44.26 & 43.36 & 45.21 \\
& $k = 2$ & 45.41 & 43.99 & 44.59 & 43.41 & 45.84 & 46.31 & 45.38 \\
& $k = 3$ & \textbf{47.98} & \textbf{46.35} & \textbf{47.69} & \textbf{45.10} & \textbf{48.72} & \textbf{49.52} & \textbf{47.96} \\
& $k = 4$ & 47.57 & 45.42 & 46.51 & 44.39 & 48.22 & 48.96 & 47.52 \\
& $k = 5$ & 46.86 & 42.82 & 42.21 & 43.45 & 46.95 & 47.08 & 46.84 \\
\midrule
\multirow{15}{*}{\rotatebox{90}{\textbf{KD-COT Module (Mental Llama)}}} 
& \multicolumn{8}{l}{\textit{$+$ve and $-$ve Labels}} \\
& $k = 1$ & 45.36 & 43.25 & 44.51 & 42.07 & 46.56 & 47.85 & 45.34 \\
& $k = 2$ & 45.01 & 43.30 & 44.68 & 42.01 & 46.36 & 47.74 & 44.97 \\
& $k = 3$ & \textbf{45.97} & \textbf{43.76} & \textbf{45.19} & \textbf{42.32} & \textbf{47.21} & \textbf{48.64} & \textbf{45.88} \\
& $k = 4$ & 43.67 & 43.11 & 44.57 & 41.76 & 45.18 & 46.87 & 43.64\\
& $k = 5$ & 43.18 & 40.92 & 41.23 & 40.63 & 44.78 & 46.54 &  43.15\\
\cmidrule(lr){2-9}
& \multicolumn{8}{l}{\textit{$+$ve Labels Only}} \\
& $k = 1$ & 43.72 & 43.50 & 44.67 & 42.39 & 44.93 & 46.34 & 43.63\\
& $k = 2$ & 45.02 & 44.34 & 45.21 & 43.52 & 45.66 & 46.71 & 44.77 \\
& $k = 3$ & \textbf{45.11} & \textbf{44.56} & \textbf{45.62} & \textbf{43.56} & \textbf{45.42} & \textbf{45.98} & \textbf{44.89} \\
& $k = 4$ & 44.87 & 44.14 & 45.28 & 43.16 & 45.00 & 45.68 & 44.35\\
& $k = 5$ & 43.17 & 43.04 & 44.52 & 41.67 & 44.33 & 45.85 & 42.91 \\
\cmidrule(lr){2-9}
& \multicolumn{8}{l}{\textit{$-$ve Labels Only}} \\
& $k = 1$ & 42.96 & 42.35 & 43.56 & 41.22 & 44.30 & 45.94 & 42.78 \\
& $k = 2$ & 42.37 & 42.27 & 43.67 & 40.96 & 43.53 & 45.31 & 41.89 \\
& $k = 3$ & \textbf{43.05} & \textbf{42.87} & \textbf{44.02} & \textbf{41.78}& \textbf{43.38} & \textbf{46.05} & \textbf{42.84} \\
& $k = 4$ & 42.75 & 41.99 & 42.66 & 41.36 & 43.94 & 45.34 & 42.63\\
& $k = 5$ & 42.63 & 41.67 & 42.83 & 40.59 & 43.81 & 45.13 & 42.57 \\
\bottomrule
\end{tabular}
}
\caption{Ablation Study for Relevant Context and KD-COT Modules. The best-performing configuration is highlighted.}
\label{tab:ablation_study}
\end{table}

\begin{figure}[t]
    \centering
    \includegraphics[width=\columnwidth]{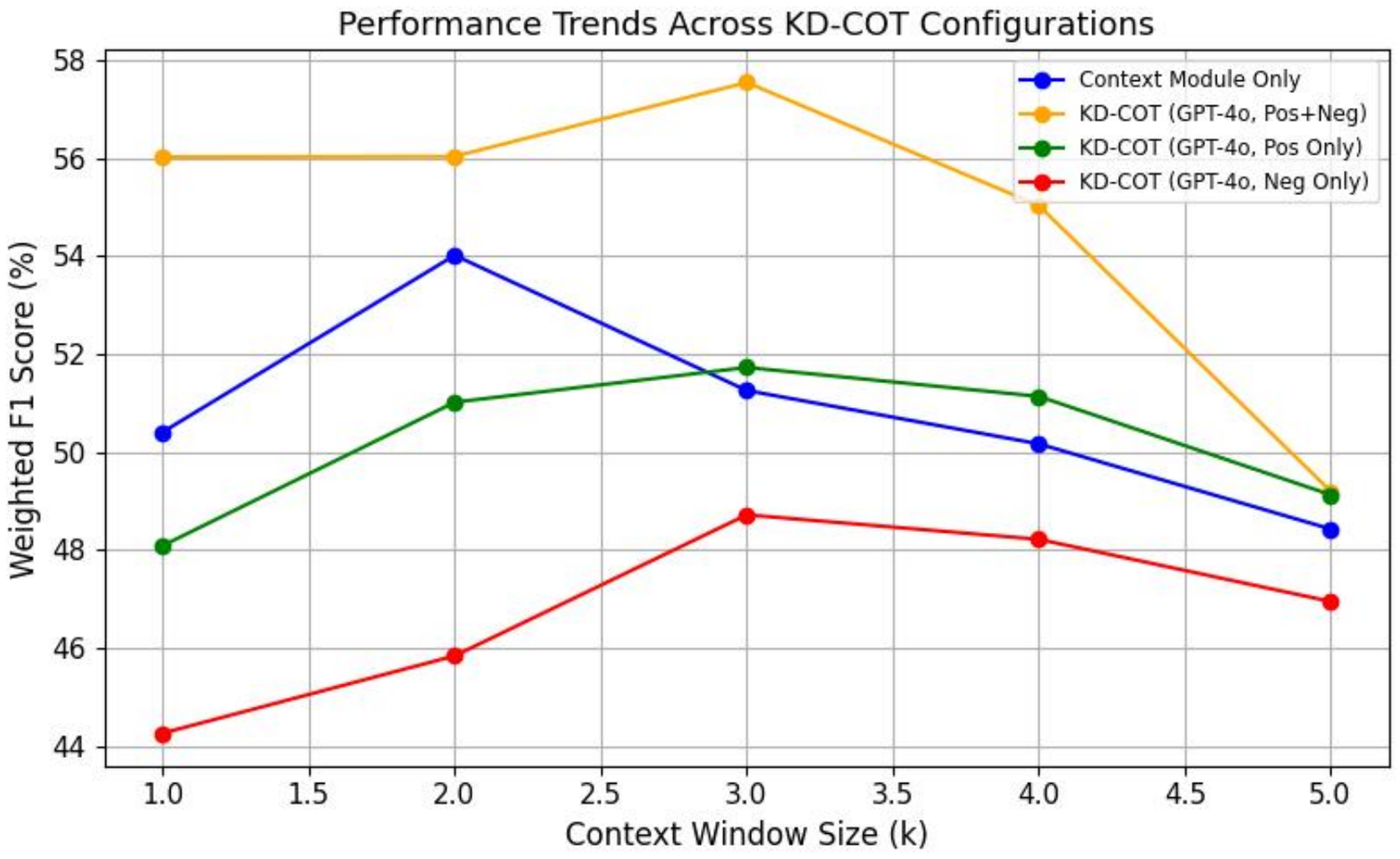}
    \caption{ Performance trends across KD-CoT configurations, showing the variation in Weighted F1 score with different context window sizes and exemplar retrieval. }
    \label{fig:kd_cot_ablation}
\end{figure}

\begin{table}[t]
\centering
\resizebox{\columnwidth}{!}{
\begin{tabular}{lccc}
    \toprule
    \textbf{Therapeutic Dimension} & \textbf{Prec} & \textbf{Rec} & \textbf{F1} \\
    \midrule
    Non-Judgmental Language        & 65.80 & 64.45 & 65.10 \\
    Warmth and Encouragement       & 62.10 & 61.70 & 61.90 \\
    Active Listening               & 64.20 & 63.40 & 63.80 \\
    Reflecting Feelings            & 62.40 & 62.00 & 62.20 \\
    Situational Appropriateness    & 63.90 & 63.30 & 63.60 \\
    Respect for Autonomy           & 63.70 & 63.20 & 63.44 \\
    \bottomrule
\end{tabular}
}
\caption{Label-wise precision, recall, and F1 scores (\%) for each therapeutic dimension on \dataset\ using \model. Highest F1 is observed for Non-Judgmental Language; lowest for Warmth and Reflecting Feelings.}
\label{tab:labelWiseResults}
\vspace{-3mm}
\end{table}

\section{Additional Ablation Details.} 
\label{sec:additionalablation}
Table~\ref{tab:ablation_study} provides a comprehensive breakdown of the ablation experiments investigating the impact of context window size (\(k\)) and the label polarity settings on the performance of the Relevant Context and KD-CoT modules. Consistent with observations discussed earlier, the Relevant Context Module achieves optimal performance at \(k=2\), balancing sufficient conversational history without introducing noise. For the KD-CoT module, evaluations across different configurations utilizing both positive and negative labels, positive only, and negative-only exemplars demonstrate that contrastive reasoning with both polarities significantly enhances accuracy and F1 scores. Notably, GPT-4o outperforms Mental LLaMA by a substantial margin across all metrics and configurations, reinforcing the critical role of a high-quality reasoning backbone. The detailed metrics, including accuracy, macro and weighted F1, precision, and recall, further validate these findings, emphasizing the robustness of the proposed framework under varying settings.

\begin{figure*}[t]
\centering
\begin{tcolorbox}[width=\textwidth, colback=gray!5, colframe=black!80, title=\textbf{Few-Shot vs Zero-Shot Prompt Design for GPT-4o Evaluation}, fonttitle=\bfseries, sharp corners=south]
\small
\textbf{Few-Shot Prompt:}

\textit{Instruction}: You are a clinical NLP evaluator. Score a therapist response from \textbf{-2 to +2} across six therapeutic principles.

You will be shown few-shot examples and then asked to score a new interaction.

\texttt{\#\#\# Example} \\
\texttt{Patient: <previous utterance>} \\
\texttt{Therapist: <therapist response>} \\
\texttt{Explanations:}
\begin{itemize}[noitemsep,nolistsep]
    \item[--] \texttt{Non-Judgmental Language: ...}
    \item[--] \texttt{Warmth and Encouragement: ...}
    \item[--] \texttt{Respect for Autonomy: ...}
    \item[--] \texttt{Active Listening: ...}
    \item[--] \texttt{Reflecting Feelings: ...}
    \item[--] \texttt{Situational Appropriateness: ...}
\end{itemize}
\texttt{Scores: NJ: X, W: X, RA: X, AL: X, RF: X, SA: X} \\

Repeat for multiple examples.

\texttt{\#\#\# Now evaluate this response:} \\
\texttt{Patient: <current patient utterance>} \\
\texttt{Therapist: <current therapist response>} \\
\texttt{Explanations: ...} \\
\texttt{Return scores in same format.}

\vspace{3mm}
\textbf{Zero-Shot Prompt:}

\textit{Instruction}: You are a mental health evaluation assistant trained in psychotherapy communication techniques.
Please assess the quality of the \textbf{therapist's response} using the following six therapeutic dimensions. Use the preceding \textbf{patient's statement} to help you understand the context and appropriateness of the therapist's message. \\ 

\texttt{Patient said: "<patient utterance>"} \\
\texttt{Therapist responded: "<therapist utterance>"} \\

Rate the therapist's response from \textbf{-2 to +2} for each dimension:
- -2 = strongly violates the principle  
-  0 = neutral or unclear  
- +2 = strongly demonstrates the principle   \\

\textbf{Output Format (verbatim):} \\
\texttt{Non-Judgmental: X, Warmth: X, Respect: X, Active Listening: X, Reflecting Feelings: X, Situational Appropriateness: X}
\end{tcolorbox}
\caption{\mm{Comparison of the few-shot (top) and zero-shot (bottom) prompt formats used to evaluate GPT-4o and Mental LLaMA responses. The few-shot prompt includes in-context examples and explanations, while the zero-shot prompt relies only on task instruction and current dialogue context.}}
\label{fig:few_zero_prompts}
\end{figure*}
\begin{figure*}[h!]
\centering

\begin{tcolorbox}[width=\textwidth, colback=gray!5, colframe=black!80, title= Prompt Design for Dimension Specific COT Explanation]\small
\textbf{Instruction}: I need you to think like a mental health counselor and use a hidden chain-of-thought process. I will provide you with a therapist dialogue along with several reference dialogues. First, internally analyze and understand the reference dialogues to build expert knowledge. Then, using your hidden chain-of-thought reasoning, analyze the therapist dialogue and develop a detailed explanation for Active Listening. Finally, provide your explanation in exactly 100 words. Do not include any of your internal reasoning in the final output. \\

\textbf{Label Exclusive Utterances}: It's really important for us to consider that kind of thinking, hmm. Okay, Marsha, it's good that we've got those things on the board. And what I'd like to do now is focus, stepping out a bit from your situation and looking, looking at it from another point of view. To me, it sounds like you've, you're doing a whole lot of things really well. Can you list all of the things that you're doing really well? \\

\textbf{Therapist Current Utterance}: Okay, you're doing great, just keep going.
\end{tcolorbox}
\caption{Prompt used to extract dimension-specific Chain-of-Thought explanations from GPT-4o. The retrieved exemplars from the previous step are passed to the model along with the structured instruction, guiding the explanation of a therapeutic principle such as Active Listening.}
\label{fig:explanation_prompt}
\end{figure*}

\begin{figure*}[t]
\centering
\begin{tcolorbox}[
    width=\textwidth,
    colback=gray!5,
    colframe=black!80,title= Model-generated Rationale for Active Listening
]
\small
\textbf{Therapist Response:}

“I appreciate you answering these questions. For me, this has been helpful.”

\vspace{0.5em}

\textbf{Model-Generated Explanation (Active Listening):}

Active listening is a crucial therapeutic skill that involves fully engaging with a client's words, emotions, and underlying meanings. In the therapist dialogue, the phrase “I appreciate you answering these questions” demonstrates validation, showing the client their input is valued. By stating, “This has been helpful,” the therapist reinforces the significance of the dialogue, encouraging further sharing. This approach fosters a safe space for clients to express themselves openly. Active listening not only enhances understanding but also builds rapport, allowing clients to feel heard and supported in their journey towards healing and self-discovery.
\end{tcolorbox}
\caption{\mm{Representative example of a model-generated rationale used within the KD-CoT pipeline for the \textit{Active Listening} therapeutic principle.}}
\label{fig:human_rationale_example}
\end{figure*}

\begin{figure*}[h]
\centering
\begin{tcolorbox}[width=\textwidth,
    colback=gray!5,
    colframe=black!80,
    title=\textbf{Full Few-Shot In-Context Example Used for GPT-4o Evaluation},
    fonttitle=\bfseries,
    sharp corners=south
]
\small
\textbf{Instruction:}  
You are an expert in mental health counselling. Evaluate each therapist's response on a scale from \textbf{-2 to +2} based on the following six therapeutic principles:

Non-Judgmental Language;  
Warmth and Encouragement;  
Respect for Autonomy;  
Active Listening;  
Reflecting Feelings;  
Situational Appropriateness.

Use the reference examples below to guide your judgment. Each example includes a therapist response, detailed explanations for each principle, and final scores. Follow the same reasoning pattern when evaluating the new response.\\

\textbf{Example A}

\textit{Therapist Response:}  
“So you have difficulty falling asleep. But once you fall asleep.”

\textit{Explanations:}  
\textbf{Non-Judgmental Language:} Non-judgmental language in therapy is essential for creating a safe and supportive environment where clients feel comfortable sharing their experiences without fear of criticism.  
\textbf{Warmth and Encouragement:} Warmth and encouragement in therapy create a safe space for clients to explore their feelings and challenges.  
\textbf{Respect for Autonomy:} Respect for autonomy emphasizes recognizing the client’s ability to make their own decisions regarding their mental health.  
\textbf{Active Listening:} Active listening involves fully concentrating on the client’s words and demonstrating understanding.  
\textbf{Reflecting Feelings:} Reflecting feelings acknowledges the client’s emotional experience by summarizing their struggle with sleep difficulties.  
\textbf{Situational Appropriateness:} The therapist response appropriately addresses a common clinical concern raised by the client.

\textit{Scores:}  
Non-Judgmental: 2, Warmth: 1, Respect: 1, Active: 2, Reflecting: 1, Situational: 2

\vspace{2mm}
\textbf{Example B}

\textit{Therapist Response:}  
“What County are we in?”

\textit{Explanations:}  
\textbf{Non-Judgmental Language:} The response does not express judgment and maintains neutrality.  
\textbf{Warmth and Encouragement:} The question provides limited warmth and encouragement.  
\textbf{Respect for Autonomy:} The response does not restrict the client’s autonomy.  
\textbf{Active Listening:} The response demonstrates minimal engagement with the client’s expressed concerns.  
\textbf{Reflecting Feelings:} The question serves as a grounding technique rather than emotional reflection.  
\textbf{Situational Appropriateness:} The response is weakly appropriate as it diverts attention from the client’s emotional needs.

\textit{Scores:}  
Non-Judgmental: 1, Warmth: 0, Respect: 1, Active: 1, Reflecting: 1, Situational: 1

\vspace{2mm}
\textbf{Now evaluate this response:}

\textit{Therapist Response:}  
“Okay, good. I’ll see you next week. And we’ll see how you did with those goals.”

\textit{Explanations:}  
\textbf{Non-Judgmental Language:} Non-judgmental language fosters a supportive therapeutic environment. The phrase “we’ll see how you did with those goals” avoids explicit criticism but implies evaluation.  
\textbf{Warmth and Encouragement:} The response conveys mild encouragement through continuity and reassurance.  
\textbf{Respect for Autonomy:} The therapist acknowledges the client’s responsibility for goal progress without imposing directives.  
\textbf{Active Listening:} The response reflects acknowledgment of prior discussion about goals and future follow-up.  
\textbf{Reflecting Feelings:} Emotional reflection is limited, as the response focuses on task completion rather than affect.  
\textbf{Situational Appropriateness:} The response is appropriate as a session-closing statement that reinforces accountability and continuity.

\end{tcolorbox}
\caption{\mm{Complete few-shot in-context example used for GPT-4o evaluation. The prompt includes prior therapist responses with full principle-wise explanations and ordinal scores, followed by a new therapist response evaluated using the same structure.}}
\label{fig:fewshot_full_example}
\end{figure*}

\section{Prompting}
\subsection{Few-shot vs Zero-shot}
The few-shot setting provides multiple in-context examples with detailed explanations, enabling the model to learn implicit scoring patterns and align its judgments accordingly. In contrast, the zero-shot prompt relies solely on task instructions and the immediate dialogue context, without prior demonstrations. This distinction highlights how contextual grounding through examples can influence model reasoning, consistency, and interpretability in therapeutic assessment tasks. Overall, Figure \ref{fig:few_zero_prompts} emphasizes the trade-off between guided learning in few-shot prompting and the generalization capability of zero-shot evaluation.

\subsection{Prompting with Label-Guided Reasoning}

\label{sec:promptsandexamples}
To obtain precise and contextually grounded explanations of therapeutic principles, we designed a structured prompt that incorporates \textit{label-exclusive} utterances dialogue snippets characteristic of a specific therapeutic dimension to guide the model’s reasoning. The prompt instructs the language model to act as a mental health counselor, engaging in internal chain-of-thought reasoning before generating a final explanation.

Specifically, the model is provided with a set of reference utterances and a target therapist statement. It is asked to internally synthesize insights from the references and produce a 100-word explanation focused on a single therapeutic principle such as \textit{Active Listening} while omitting any visible reasoning steps. This approach encourages the model to generate concise yet rich rationales grounded in domain-relevant context.

An example of the full prompt template used to elicit such explanations from GPT-4o is shown in Figure~\ref{fig:explanation_prompt}. The retrieved exemplars from the previous stage (i.e., label-exclusive utterances) are incorporated into the prompt, which guides the model in producing dimension specific justifications aligned with therapeutic principle.

\section{Human-Level Rational Assessment}
\label{sec:human_rational_assessment}

\mm{CARE incorporates expert-level reasoning through the KD-CoT module, which relies on large language models to generate fine-grained, expert-standard explanations that guide ordinal classification. A natural concern arises regarding the credibility of these generated rationales, particularly in light of the strong agreement observed between expert annotations and CARE across multiple dimensions in Table \ref{tab:expert_agreement}.}

\mm{To address this concern, we conduct a human-level explanation assessment to evaluate the quality of rationales generated by GPT-4o and Mental-LLaMA, which are subsequently utilized within the CARE framework. We randomly selected 15 instances, and for each instance, three clinically relevant dimensions were chosen. For each dimension, both LLMs generated an explanation.}

\mm{These explanations were independently evaluated by 25 human evaluators. Each explanation was rated on a 5-point Likert scale (1 = poor, 5 = excellent), focusing on clarity, contextual relevance, and clinical appropriateness.}

\mm{Our analysis indicates that GPT-4o explanations are consistently rated higher, with approximately 80\% receiving a score of 4 or 5, reflecting strong alignment with expert reasoning and contextual grounding. In contrast, Mental-LLaMA explanations are predominantly rated between 3 and 4, suggesting moderate quality with occasional limitations in specificity or contextual depth. These findings suggest that GPT-4o produces more reliable and clinically coherent rationales, making it a strong choice for guiding CARE through the KD-CoT mechanism.}

\paragraph{Rationale Examples}
\mm{Figure~\ref{fig:human_rationale_example} presents a representative example of a model-generated rationale for the \textit{Active Listening} dimension. The explanation highlights clinically meaningful cues such as validation, responsiveness, and conversational engagement. These examples are intended to demonstrate the clarity, contextual relevance, and clinical plausibility of the generated rationales, rather than to serve as ground-truth therapeutic justifications.}

\begin{figure*}[t]
    \centering
    \includegraphics[width=\textwidth]{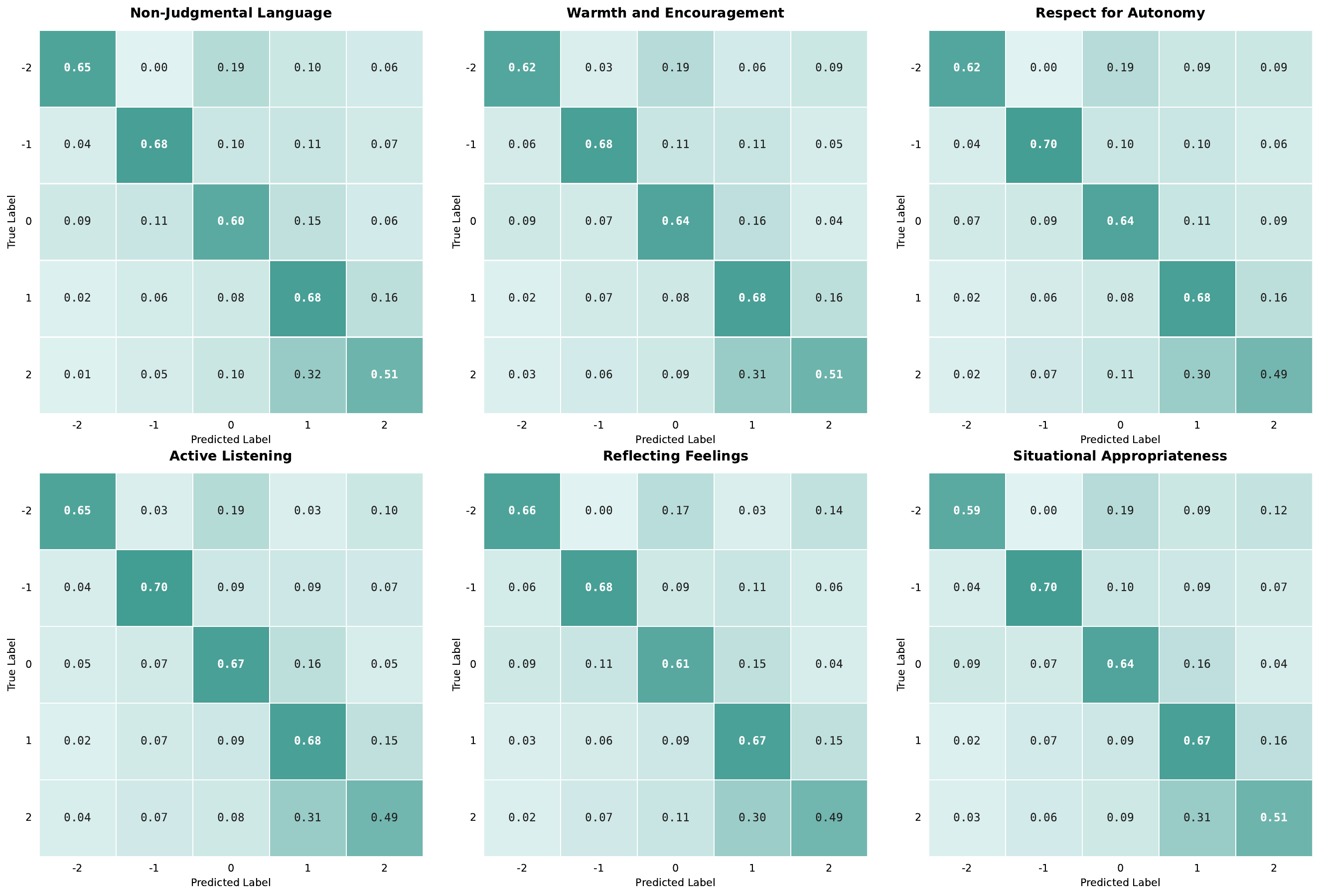}
    \caption{Confusion matrices computed on our proposed dataset, \dataset\ using our model \model, across six therapeutic principles. 
    }
    \label{fig:confusionmatrix}
\end{figure*}

\begin{figure*}[t]
    \centering
    \includegraphics[width=\textwidth]{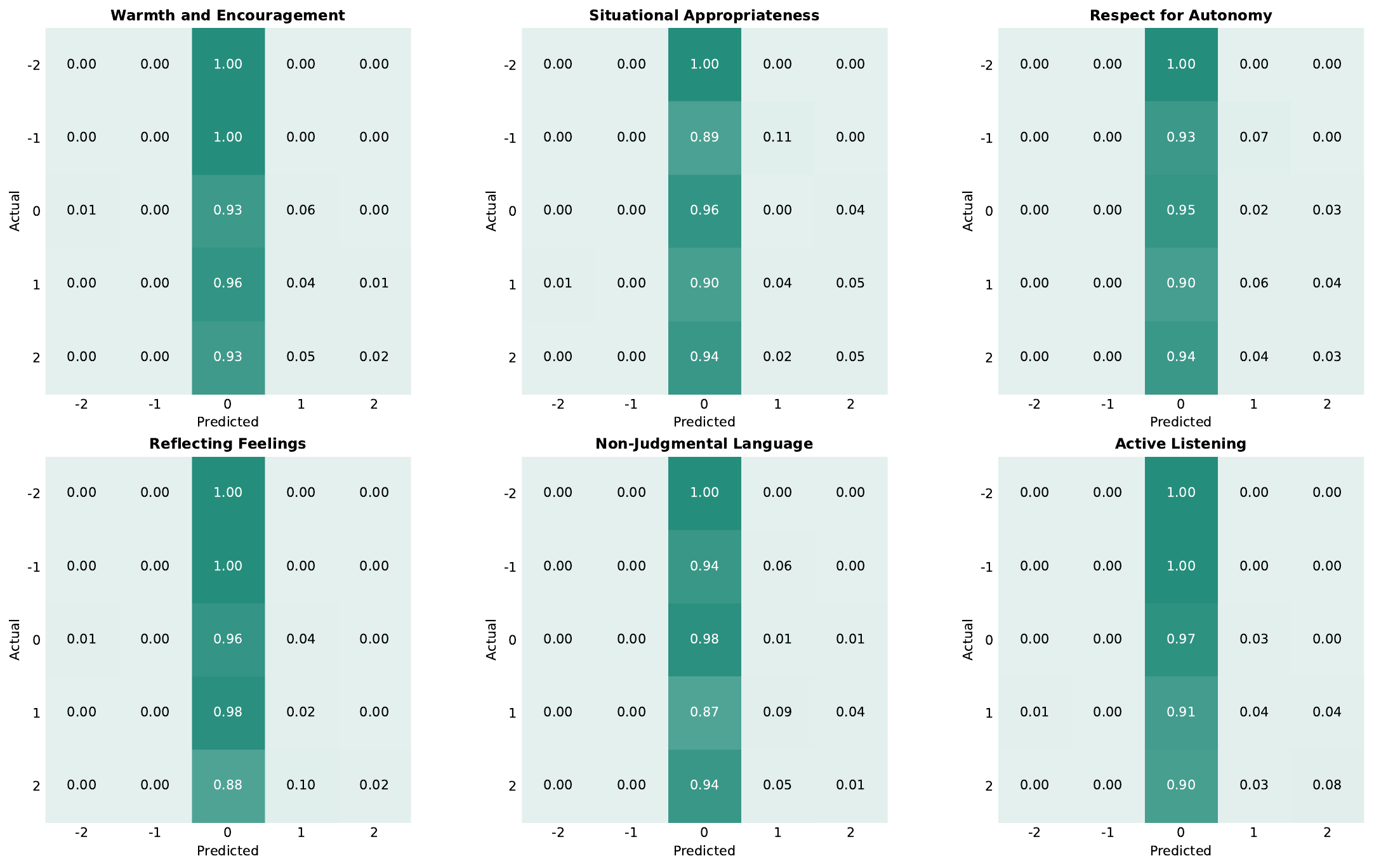}
    \caption{\mm{Confusion matrices computed on \dataset\ using GPT-4o under zero-shot prompting across six therapeutic principles. The matrices reveal a strong bias toward positive predictions, with negative and neutral categories frequently collapsed into Mild Positive or Strong Positive classes.}}
    \label{fig:gpt_zero_cf}
\end{figure*}

\begin{figure*}[t]
    \centering
    \includegraphics[width=\textwidth]{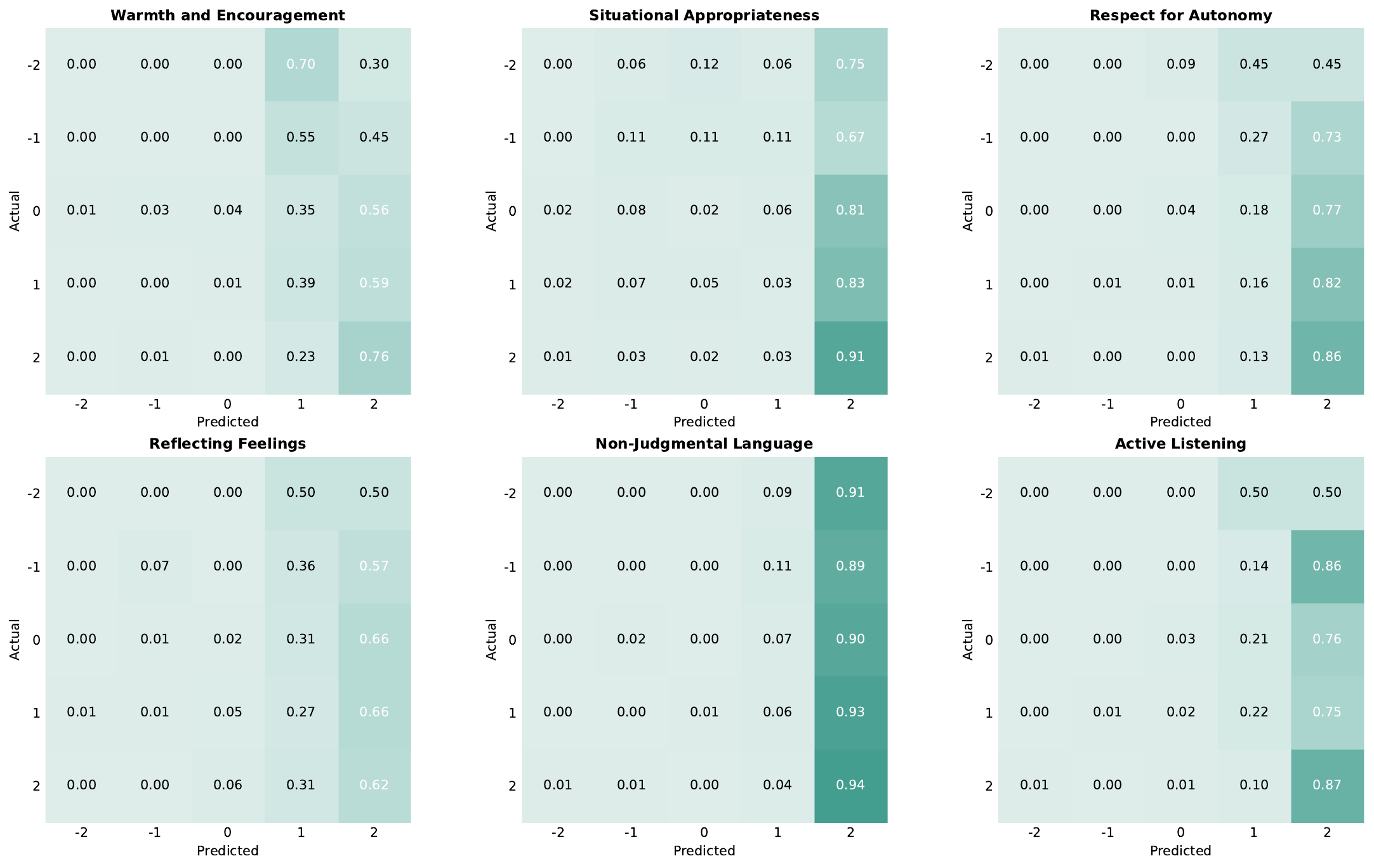}
    \caption{\mm{Confusion matrices computed on \dataset\ using GPT-4o under few-shot prompting across six therapeutic principles. While few-shot prompting sharpens confidence around positive labels, it does not correct the systematic under-recognition of negative and neutral categories.}}
    \label{fig:gpt_few_cf}
\end{figure*}

\section{Behavioral Differences Between CARE and GPT-4o}

Figure~\ref{fig:confusionmatrix} presents normalized confusion matrices across the six therapeutic dimensions, comparing CARE with GPT-4o under zero-shot and few-shot prompting. Rows denote gold labels and columns predicted labels under the ordinal scale: SN, MN, N, MP, and SP.

Across dimensions, CARE exhibits a coherent ordinal error structure, with most misclassifications occurring between adjacent categories (e.g., Neutral vs.\ Mild Positive). This behavior indicates sensitivity to ordinal severity and reliable discrimination at the extremes (SN and SP), reflecting robust handling of clearly misaligned and strongly aligned responses.

In contrast, GPT-4o under zero-shot prompting (Figure~\ref{fig:gpt_zero_cf}) shows a strong bias toward Mild Positive and Strong Positive predictions, frequently collapsing negative and neutral instances into positive categories. This effect is particularly evident for principles such as \textit{Non-Judgmental Language} and \textit{Respect for Autonomy}, suggesting reliance on surface-level linguistic cues rather than relational or pragmatic intent.

Few-shot prompting improves output consistency but does not correct this bias. As shown in Figure~\ref{fig:gpt_few_cf}, predictions remain skewed toward positive labels, with limited recovery of negative and neutral classes. Compared to CARE, GPT-4o’s errors are less ordinally localized, highlighting the limitations of prompt-based evaluation for nuanced therapeutic assessment.

\end{document}